\documentclass[10pt,letterpaper]{article}
\usepackage[top=0.85in,left=2.75in,footskip=0.75in]{geometry}

\usepackage{amsmath,amssymb}

\usepackage{changepage}

\usepackage[utf8x]{inputenc}

\usepackage{textcomp,marvosym}

\usepackage{cite}

\usepackage{nameref,hyperref}

\usepackage[right]{lineno}

\usepackage{microtype}
\DisableLigatures[f]{encoding = *, family = * }

\usepackage{xcolor}
\usepackage{array}

\newcolumntype{+}{!{\vrule width 2pt}}

\newlength\savedwidth

\raggedright
\usepackage[aboveskip=1pt,labelfont=bf,labelsep=period,justification=raggedright,singlelinecheck=off]{caption}

\makeatletter
\renewcommand{\@biblabel}[1]{\quad#1.}
\makeatother

\usepackage{lastpage,fancyhdr,graphicx}
\usepackage{epstopdf}
\fancyheadoffset[L]{2.25in}
\fancyfootoffset[L]{2.25in}
\begin{document}
\vspace*{0.2in}

\begin{flushleft}
{\Large
\textbf\newline{Pre-Synaptic Pool Modification (PSPM): A Supervised Learning Procedure for Recurrent Spiking Neural Networks} 
}
\newline
\\
Bryce Bagley\textsuperscript{1,2,3,4 *},
Blake Bordelon\textsuperscript{1,2},
Benjamin Moseley\textsuperscript{3,5},
Ralf Wessel\textsuperscript{2},
\\
\bigskip
\textbf{1} Department of Electrical and Systems Engineering, Washington University in St. Louis, St. Louis, MO, USA
\\
\textbf{2} Department of Physics, Washington University in St. Louis, St. Louis, MO, USA
\\
\textbf{3} Department of Computer Science, Washington University in St. Louis, St. Louis, MO, USA
\\
\textbf{4} Stanford Institute for Theoretical Physics, Stanford University, Stanford, CA, USA
\\
\textbf{5} Department of Operations Research, Carnegie Mellon University, Pittsburgh, PA, USA
\\
\bigskip

%
%

* bbagley@stanford.edu

\title{Pre-Synaptic Pool Modification (PSPM): A Supervised Learning Procedure for Recurrent Spiking Neural Networks}



\author{Bryce Bagley}
\author{Blake Bordelon}
\author{Benjamin Moseley}
\author{Ralf Wessel}

\end{flushleft}
\section*{Abstract}
Learning synaptic weights of spiking neural network (SNN) models that can reproduce target spike trains from provided neural firing data is a central problem in computational neuroscience and spike-based computing. The discovery of the optimal weight values can be posed as a supervised learning task wherein the weights of the model network are chosen to maximize the similarity between the target spike trains and the model outputs. It is still largely unknown whether optimizing spike train similarity of highly recurrent SNNs produces weight matrices similar to those of the ground truth model. To this end, we propose flexible heuristic supervised learning rules, termed Pre-Synaptic Pool Modification (PSPM), that rely on stochastic weight updates in order to produce spikes within a short window of the desired times and eliminate spikes outside of this window. PSPM improves spike train similarity for all-to-all SNNs and makes no assumption about the post-synaptic potential of the neurons or the structure of the network since no gradients are required. We test whether optimizing for spike train similarity entails the discovery of accurate weights and explore the relative contributions of local and homeostatic weight updates. Although PSPM improves similarity between spike trains, the learned weights often differ from the weights of the ground truth model, implying that connectome inference from spike data may require additional constraints on connectivity statistics. We also find that spike train similarity is sensitive to local updates, but other measures of network activity such as avalanche distributions, can be learned through synaptic homeostasis.


\section*{Introduction}
\label{S:1}

With the advent of high dimensional multi-electrode and calcium imaging recordings of neural activity, efficient computational methods are required to discover the underlying operational principles of neural microcircuits \cite{MAGRAN}. Such computational tools would allow the inference of connectivity structures at the microcircuit scale from experimental recordings \cite{SpornsConnectome}. Although an analysis of functional connectivity, or statistical dependence of spiking behavior in neurons, requires only descriptive statistics of the provided spike data, a generative model of causal connectivity can be obtained by fitting the weights of model networks to the experimental data \cite{Friston}.

Two considerations predominantly drive the decision of which model of neuron dynamics to use in such an inference task: biological plausibility and computational efficiency. One of the most biologically plausible model of the membrane potential dynamics is given by the Hodgkin and Huxley equations \cite{Hodgkin}, which was developed as a dynamical model of the of the squid axon. Each neuron in the Hodgkin and Huxley model consists of four coupled first order differential equations. While Hodgkin and Huxley-like models accurately describe the dynamics of a neuron's membrane potential dynamics in a high level of detail, they are computationally expensive, motivating the study of simpler models that preserve the important aspects of neural dynamics. 

Spiking neural network (SNN) models, including the Spike Response model (SRM) \cite{Jolivet}, Izhikevich model \cite{Izhikevich} and Leaky Integrate and Fire (LIF) models \cite{Delorme}, are computationally much simpler than the Hodgkin and Huxley model while still maintaining biological plausibility. In addition to the computational justification for these models, several empirical studies have demonstrated the reliability of spike timing in the brain \cite{Mainen} \cite{Herikstad} \cite{Butts}, suggesting that spike timing plays an important role in the neural code and justifying models where spike times play an explicit role in the network dynamics. Thus, fitting SNNs that reproduce experimental spike trains could plausibly be used to discover the effective or causal connectivity of the network under study. This aim motivates a methodology for inferring weights of SNNs based on their spike trains. 

Training SNNs to efficiently perform computation, including machine learning tasks, also motivates the study of learning rules for SNNs \cite{maass2}. Since the spikes in SNNs are sparse in time, SNNs can be trained on provided data with considerably fewer operations, saving both energy and time when compared to their artificial neural network (ANN) counterparts \cite{Tavanaei}. SNNs have been applied to audio-visual processing \cite{WYSOSKI}, edge detection \cite{Wu}, character recognition \cite{Gupta}, and speech recognition \cite{Liaw}. Biologically plausible learning rules for spiking neural networks have also been employed for unsupervised learning tasks like visual feature extraction \cite{Masquelier}, sparse coding \cite{Tang}, and non-negative similarity matching \cite{Pehlevan}.  
The observation that SNNs can efficiently perform several machine learning tasks has motivated the development of neuromorphic hardware specially designed for simulating the dynamics of neural microcircuits \cite{Mead} \cite{IBM} \cite{nahmias}. Designing the synaptic connections in a neuromorphic system to accomplish a particular task similarly requires learning rules for SNNs, further motivating the development of an algorithmic framework for designing synaptic connections that can perform a particular task.

Work on supervised training of SNNs began with the SpikeProp algorithm, which was an analogue of backpropagation in ANNs, an algorithm that computes the gradients of a loss function $E$ with respect to the weight values \cite{bohte}. The loss function for SpikeProp compared the desired spike time with the observed spike time with a least squares objective. The post-synaptic potential kernel $\epsilon(t)$, which describes the impact of synaptic currents on the membrane potential of the post-synaptic neuron, was differentiable, permitting the calculation of gradients of the SpikeProp loss function $\frac{\partial E}{\partial w}$ with the chain rule. Using gradient descent, SpikeProp was capable of training feed-forward SNNs for the XOR problem, but it was limited in that it only allowed one spike per neuron, was very sensitive to the initialization of weight parameters, and demanded latency based coding. Other gradient based algorithms relaxed these restrictions by allowing multiple spikes per neuron \cite{booij}, multiple neurons \cite{GHOSHDASTIDAR20091419}, and allowing the synaptic delays and time constants to be free parameters in the optimization problem \cite{schrauwen1} \cite{schrauwen2}. Backpropagation takes advantage of layered, feedforward structure to efficiently compute gradients with the chain rule. In networks with a high degree of recurrence or all-to-all connectivity, other strategies must be employed. 

Learning rules for a probabilistic SNN model have been developed by Pfister et. al, in which the likelihood of a spike occurring at the desired time is maximized with gradient ascent \cite{Pfister}. A similar maximum likelihood technique was applied to the fitting parameters of the Mihalas-Niebur neuron model \cite{russell}.  Although stochastic model neurons perform well for randomly exploring reinforcement learning policies, they are not ideal for spike time reproduction, in which low variability in output spike trains for fixed weights is preferred. Following this work, Gardner et. al. adapted the weight updates from Pfister's stochastic model to optimize instantaneous and filtered error signals in a deterministic LIF network \cite{gardner}. In these works, post-synaptic potentials were also required to compute gradients of the likelihood functions.

Chronotron \cite{Florian}, ReSuMe \cite{Ponulak}, and SPAN \cite{Mohemmed} all provide learning rules for a single spiking neuron receiving input from many pre-synaptic neurons. The Chronotron, like other learning algorithms, attempts to produce spikes at desired times, but rather than using using post-synaptic potentials to compute gradients of an error function, it relies on the Victor-Purpora (VP) metric between two spike trains. The VP distance measures the cost associated with transforming one spike train into another by creating, removing, or moving spikes. The Chronotron uses an adaptation to the VP distance that renders it differentiable and thus amenable to gradient descent with respect to the weights. ReSuMe (remote supervised learning) and SPAN both adapt variants of spike time dependent plasticity (STDP) and anti-STDP rules in which the weight change is proportional to the difference between the desired and observed spike trains. SPAN uses identical learning rules but filters the spike trains with the alpha kernel $\epsilon(t) \sim te^{-t / \tau}$, essentially converting the digital spike trains to an analog sum of post-synaptic potentials.  

The application of general purpose evolutionary algorithms to train SNNs has also been successfully explored \cite{jin} \cite{pavlidis}. These methods, inspired by biological evolution, explore the space of possible weight matrices and receive feedback from a loss function during training. Although these methods optimize the weights of SNNs and do not require knowledge of the post-synaptic potentials or gradients, they do not leverage the domain knowledge specific to this problem, namely that a given spike has a causal history that can be traced back to the recent spikes of other neurons.

Our PSPM learning rules provide a compromise between the stochastic evolutionary search methods mentioned above with more targeted gradient based local learning rules. PSPM works for all-to-all neural networks and does not require knowledge of the functional forms of post-synaptic potentials of neurons in the network. However, since it still leverages information about pre-synaptic neurons that fired in the recent past in order to make weight updates, it is more targeted and plausible than evolutionary search strategies.

Similar to the Chronotron's VP distance, our PSPM algorithm focuses on eliminating or inducing spikes so that all of the desired spikes have a pair in the observed spike trains. This is accomplished by optimally pairing spikes in the desired spike trains with those in the observed spike trains with a dynamic program. For each unpaired spike, all pre-synaptic neurons that fired within a window of time prior to the unpaired spike of interest have their weights stochastically increased or decreased. Our learning rules do not require knowledge of the post-synaptic potential kernel but rather only require the heuristic that the causal history of a given spike can be summarized by the firing of other spikes in the recent past. We refer to these updates made to eliminate or induce unmatched spikes as local weight changes, in contrast to non-local weight changes in which the synapse between two neurons can change due to the activity of some other neuron in the newtwork.

We balance these local weight changes with network-wide homeostatic updates \cite{turrigiano}.  Our algorithm not only makes local weight updates based on the pairing rules described above but also responds to excess or inadequate network-level activity by modifying the strength of their synapses. In response to strong and weak external inputs, this modification prevents, respectively, oversaturation and extinction of spiking activity. Synaptic scaling of this sort is crucial in biological networks with recurrence, which are otherwise at risk of runaway activity resulting from feedback loops within the network \cite{keck}.

We report two major empirical findings by training LIF networks on a set of desired spike trains. First, we find that PSPM successfully reproduces the desired spike trains. However, we also find that the learned weights may differ dramatically from the weights of the ground truth network. Our finding suggests that potentially many different weight matrices can produce the same spike trains. This is significant for the project of connectome inference from neural activity data, indicating that additional knowledge about overall connectivity statistics, like weight matrix sparsity, may be necessary to accurately infer the weights. 

In addition, we find that overall network activity is sensitive to the precise location of the local weight updates and not just to the overall number or magnitude of changes. To assess the relative contributions of Hebb and homeostasis, we introduce a control network that receives fake local updates. The purpose of this control network is to demonstrate that the PSPM rule, rather than simple bio-realistic synaptic scaling, is responsible for the observed changes. While it was not anticipated that the control network would exhibit behavior similar to the PSPM optimized network, it serves as a test of the efficacy of PSPM above and beyond homeostatic changes. For every update made in the PSPM learned network, the control network receives a change but at a \textit{random synapse}. Control network spike trains vary dramatically from those of the network trained with PSPM. Unsurprisingly, exact spike matching requires local updates at the synapse locations demanded by PSPM. More surprising, however, was that the precise location of weight updates demanded by PSPM resulted in different spike trains and network wide inter-spike-interval distributions than those of the control network. This indicates that even when homeostatic adjustments are made in the network, local learning updates dominate which network-wide activity pattern is learned. 

As an additional demonstration of PSPM and its ability to train generative models that recover interesting features in spike data, we train a LIF network with spike trains provided by a critical probabilistic integrate and fire model. While spike train similarity measures were maximized with the precise weight updates of PSPM, we find that the control condition, which has weight updates at random synapses, can also change the collective firing patterns of neurons closer to the critical regime. This indicates that tuning a network to a near critical state may simply require homeostatic adjustments, whereas learning precise spike trains requires reasonable local learning rules. 

It is worth noting that our algorithm is general-purpose enough to optimize networks which have either or both of the properties of strong spike-timing-dependence and recurrence. The former is most distinct in the peripheral nervous system, and the latter in the central nervous system, with presumably some intersections. This flexibility is a key strength of PSPM, as most other algorithms cannot handle both spike-timing-dependence and recurrence.\cite{segundo}\cite{bryant}\cite{dayhoff}\cite{abeles}\cite{villa} 

While we do not analyze this hypothesis in this paper, it is worth considering the possibility that PSPM serves to produce an attractor dynamic in the optimized neural network. Given the highly nonlinear nature of spiking neural network and the observed attractors found in studies of simpler nonlinear neural network models, such dynamics are an entirely plausible explanation for the efficacy of PSPM \cite{cabessa}\cite{cabessa2}\cite{kobayashi}\cite{asai}

In summary, we present an algorithm for optimizing the weights of fully recurrent spiking neural networks. The algorithm is then tested by comparing its output to the desired spike train using a number of statistical measures such as ISI distribution similarity and the van Rossum distance metric \cite{vanrossum}. The algorithm is then further analyzed by feeding in so-called critical spike trains and using the standard metrics of the neural criticality hypothesis \cite{karimipanah2} to analyze PSPM's output. 

\section*{Methods}

\subsection*{Leaky Integrate-and-Fire (LIF) Model}

Our neural networks consisted of Leaky Integrate-and-Fire (LIF) neurons. For all neurons $i$ in the network, the membrane potential $V_i(t)$ had a time dependence given by the following equations:\\

If $V_i(t)<V_{th} $, then:

\begin{equation}
\tau \frac{dV_i(t)}{dt} = - V_i(t) + R_m I_i(t) + \sum\limits_{j} W_{ij}\ s_j(t)
\end{equation}

Else if $V_i(t)\geq V_{th}, then:$ \\
\begin{equation}
V_i(t) \gets 0 ; \ s_i(t) \gets 1 
\end{equation}

Where $\tau$ is membrane time constant, $V_{t h}$ is the voltage threshold, $R_m$ is the membrane resistance, and $I_i(t)$ is the external input current into neuron $i$ at time $t$. Further, $W_{i j}$ is the weight of the network connection from neuron $j$ to neuron $i$, and $s_j(t)$ is the binary value denoting whether or not neuron $j$ spiked at time $t$. $V_{t h}$ represents the threshold membrane potential above which a neuron will spike. In our simulations, we imposed the condition that 20\% of the neurons in our network are inhibitory with $W_{i j} < 0$ for all inhibitory neurons $j$. Network connectivity was all-to-all but we allowed $W_{i j}=0$, corresponding to an absence of synaptic strength between neuron $j$ and neuron $i$. The parameters of our neuron model are provided in Table \ref{neuron_params}. These parameters were based on typical values for biological neurons. The differential equation is evaluated discretely using Euler's method with a step size of 3 ms. \\

\begin{table}[]
    \centering
    \begin{tabular}{|c|c|c|c|c|}
        \hline
         $\tau$ & $R_m$ & $V_{th}$ & Inhibitory Synapses & step-size \\
         \hline
         30 ms & 100 M$\Omega$ & 30 mV & 20\% & 3 ms\\
         \hline 
    \end{tabular}
    \caption{Parameters used for our neuron model.}
    \label{neuron_params}
\end{table}

\subsection*{Initialized Network Parameters}
In our simulations, we used one LIF network, termed the reference network, as the ground truth model. The spike trains of the reference network are intended to be analogous to the spike trains obtained in an experimental recording or the desired outputs of a neuromorphic circuit. A distinct network called the naive network was also generated and then optimized with our learning rules to produce output spike trains similar to those of the reference. All networks consisted of $N = 400$ LIF neurons with 20\% of the neurons $j$ randomly assigned as inhibitory ($W_{ij} < 0$ for all inhibitory neurons $j$) and the remaining 80\% set as excitatory ($W_{ij} > 0$ for all excitatory neurons $j$).

Initial synaptic weights of the reference and naive networks, $W^{(R)}_{i j}$ and $W^{(N)}_{i j}$ respectively, had magnitudes that were drawn from 4 distinct distributions: (1) \textit{uniform}: reference and naive network weight matrix magnitudes were both drawn from a uniform distribution, (2) \textit{gaussian}: reference and naive weight values were both drawn from a Gaussian distribution, (3) \textit{sparse}: reference and naive weight matrix values first drawn from a uniform distribution with $50\%$ of the synapses in the naive network subsequently set to $0 \ V$, (4) \textit{naive-half-max}: reference and naive weight magnitudes drawn different uniform distributions. The distributions and initial network parameters are provided in Table \ref{initial_weight_params}. The variety of initial weight distributions was chosen to demonstrate the versatility of the algorithm for \textit{uniform} and \textit{gaussian} initializations. The \textit{sparse} and \textit{naive-half-max} initial configurations demonstrate that the algorithm can improve the performance of a naive network with initial weights drawn from a different distribution from that of the reference. 

\begin{table}[]
    \centering
    \begin{adjustwidth}{-0.70in}{0in}
    \begin{tabular}{|c|c|c|c|c|c|}
         \hline
         & N & $|W^{(R)}_{i j}|$ distribution & $|W^{(N)}_{i j}|$ distribution & Constraints\\
         \hline
         \textit{uniform} & 400 & $U[0, 5]$ mV & $U[0, 5]$ mV & \\
         \hline
         \textit{gaussian} & 400 & $\mathcal{N}(\mu = 0.4, \sigma = 0.4)$ mV & $\mathcal{N}(\mu = 0.4, \sigma = 0.4)$ mV &  $|W_{i j}| \geq 0$ \\
         \hline
         \textit{sparse} & 400 & $U[0, 5]$ mV & $U[0, 5]$ mV & 50\% of $W^{(N)}_{ij} \leftarrow 0$ \\
         \hline
         \textit{naive-half-max} & 400 & $U[0, 5]$ mV & $U[0, 2.5]$ mV & \\
         \hline
    \end{tabular}
    \end{adjustwidth}
    \caption{Initial network parameters for each of the four initial weight distributions used in this study. }
    \label{initial_weight_params}
\end{table}

The external input current values at each time step were drawn from a Gaussian distribution  $\mathcal{N}(\mu = 2.5 \times 10^{-10} A, \sigma = 1 \times 10^{-10} A)$. The values of the input currents were tuned to be strong enough to produce a network-wide mean spike frequency $0<f\leq 1$ Hz when all weights were set to zero, but could produce mean spike rates greater than 10 Hz when the weights were nonzero. This restriction guaranteed that the external input currents remained smaller than the currents generated by synaptic connections within the network, and thus the network weights significantly affected the output spike trains. The distribution described above met these requirements. \\

\subsection*{Learning Paradigm}
\begin{figure}
	\centering
		\includegraphics[width = .6\textwidth]{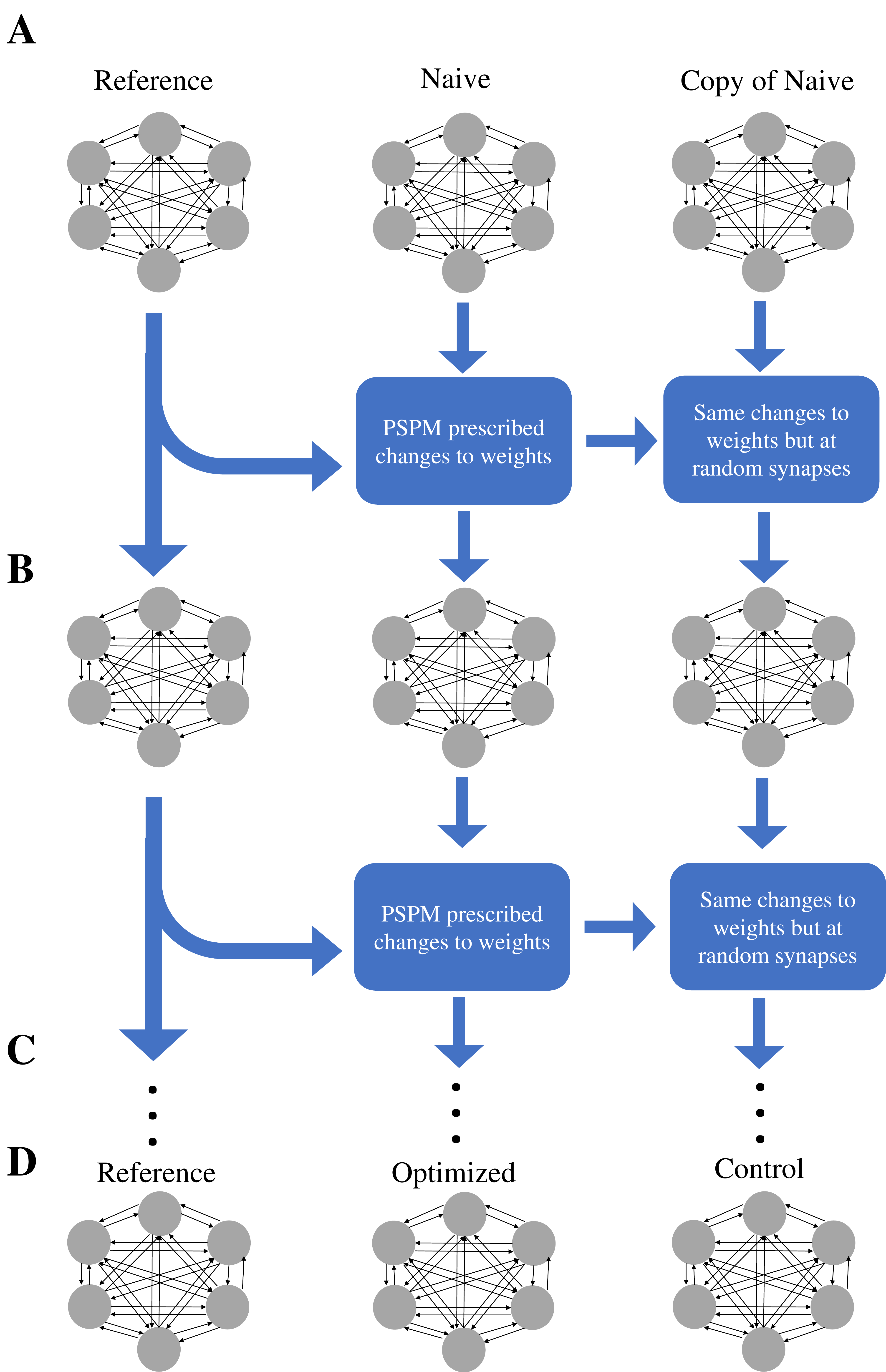}
	\caption{Learning Paradigm (\textbf{A}) Three networks are initialized: the reference network, a naive network and its copy. In addition, a set of input currents is generated and will be used throughout the trial. When simulated, the reference network produces goal spike trains $\mathcal{R}$ that the learning rule aims to reproduce in the optimized network. A naive network and its identical copy are created with weights distinct from the reference. (\textbf{B}) The reference and naive networks are simulated with the same input currents to generate outputs $\mathcal{R}$ and $\mathcal{N}$ and the PSPM learning rules are followed to change the weights of the naive network, henceforth denoted as the pre-optimized network (middle column). For every change in the pre-optimized network, there is an identical change in the copy, except the change is made at a random synapse. (\textbf{C}) This process is repeated for the specified number of epochs (150) of the algorithm. (\textbf{D}) Upon completion, the network in the middle column is considered optimized, the reference (left column) is unchanged and the remaining network (right column) is considered the control.}
	\label{learning_paradigm}
\end{figure}

We conducted 30 trials for each of the 4 sets of initial network parameters described above (\textit{Initial Network Parameters}), creating a reference network and a naive network by drawing weights from the specified distributions. The reference and naive networks are simulated with the same input currents to generate reference outputs $\mathcal{R}$ and naive outputs $\mathcal{N}$ respectively. The naive network's weights are then changed according to the PSPM rules (\textit{PSPM Learning Rules}). This modified network is simulated again with the same inputs to produce new observed spike trains $\mathcal{O}$, whose dissimilarity with $\mathcal{R}$ determines a new set of PSPM prescribed weight changes. This process is repeated for 150 epochs resulting in a network we define as optimized (Figure \ref{learning_paradigm}) with output spike trains $\mathcal{O}$ that are now considered optimized. Note that, henceforward, we will use the script notation $\mathcal{R}$, $\mathcal{N}$, $\mathcal{O}$, $\mathcal{C}$ to refer to the spike trains of the reference, naive, optimized, and control networks respectively and lowercase $r_i(t)$, $n_i(t)$, $o_i(t)$, and $c_i(t)$ to refer to the the spike train of the $i$th neuron in each of these network outputs.

\indent
To evaluate whether improvements in similarity between $\mathcal{R}$ and $\mathcal{O}$ (\textit{Spike Train Similarity Measures}) are not solely due to network-wide homeostatic adjustments to the weights but rather are due to the precise, synapse-by-synapse weight changes specified by our PSPM method, a control condition was established. At the beginning of the learning procedure, the control network is instantiated as a copy of the naive network. During the course of optimization, changes were made to the control network's weights according to the following rule: for every synaptic weight change in the naive network, an identical change is made in the control network but at a \textit{random synapse} (Figure 1). Thus the control network was provided ``fake" local updates that preserved the total number and magnitude of weight changes to the network but placed them at the wrong synapses. After the 150 epochs of training, this network is defined to be the control and its output spike trains defined as $\mathcal{C}$. Because the control network benefited from the same number and magnitudes of weight changes that the optimized network did, any difference in performance between the optimized and control networks is due entirely to the precisely specified weight alterations called for by our PSPM learning rules. \\

\begin{figure}
	\centering
		\includegraphics[width = .8\textwidth]{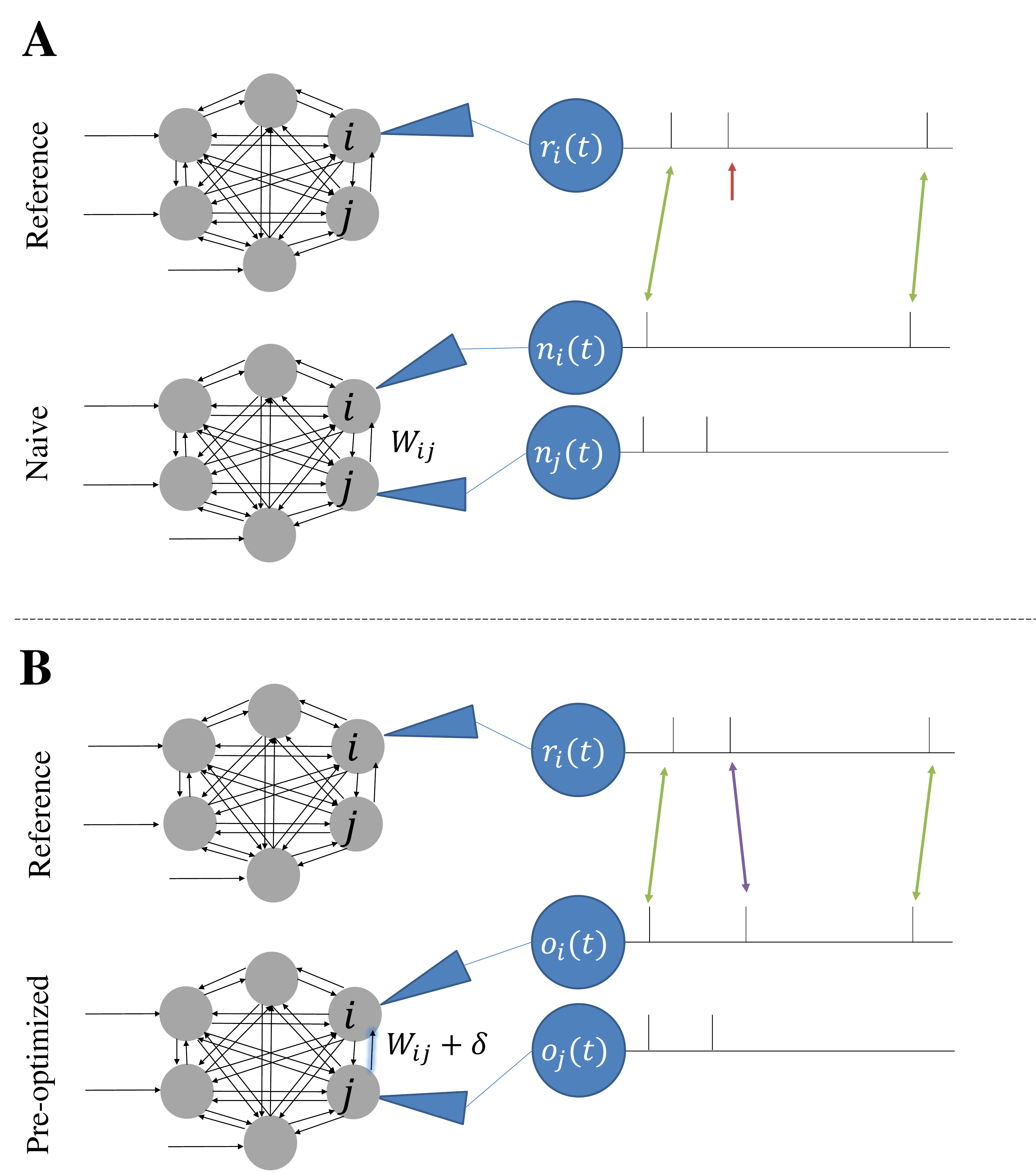}
	\caption{Pre-synaptic Pool Modification (PSPM) Learning Rules. (\textbf{A}) Spike trains for the $i_{th}$ neuron from the reference outputs $r_i(t)$ and the naive network outputs $n_i(t)$ before weight alteration. The $j_{th}$ neuron’s spike train is also shown for the naïve net $n_j(t)$. The red arrow identifies a spike in the reference output which is initially unpaired. (\textbf{B}) Because the $j_{th}$ neuron of the naïve network experiences a spike at a time shortly before the unpaired spike in the reference spike train, the synaptic weight from $j$ to $i$ is increased (blue arrow), ideally resulting in a new spike in the $i_{th}$ neuron's spike train. In the next simulation round, this new spike matches with the formerly unpaired reference spike, as indicated by the purple arrow. Networks receive this treatment for every spike that remains unpaired after application of the dynamic program (Methods).}
	\label{PSPM_algorithm_figure}
\end{figure}

\subsection*{Pre-Synaptic Pool Modification (PSPM) Learning Rules}

PSPM learning rules focus on the induction and elimination of spikes so that every desired spike has a counterpart in the produced spike train within a small temporal interval. At a high level, our algorithm consists of stochastic local weight updates aimed eliminating unmatched spikes or inducing spikes at the desired time. Additional weight modifications mimic synaptic homeostasis to keep the network at a desired level of activity. In response to differences between $\mathcal{R}$ and $\mathcal{O}$, individual weight values are additively modified (Figure \ref{PSPM_algorithm_figure}).  

To assess how well the spikes in the reference outputs match up with those of the observed outputs, we optimally pair individual spikes in the $i$th observed spike train $o_i(t)$ with their counterparts in the $i$th reference spike train $r_i(t)$ using a string-matching dynamic program. This dynamic program begins by creating a cost matrix $\Lambda \in \mathbb{R}^{n \times m}$ with elements that represent the cost of matching each possible spike pair where $n$ is the number of spikes in $r_i(t)$ and $m$ is the number of spikes in $o_i(t)$. $\Lambda_{k l}$, therefore, represents the cost of optimal pairing of the first $k$ spikes of $r_i(t)$ to the first $l$ spikes of $o_i(t)$. Note that this optimal pairing could leave many spikes unpaired as shown in Figure \ref{PSPM_algorithm_figure}.

We define the absolute temporal difference $d_{k, l}$ between each spike $k$ in $r_i(t)$ and spike $l$ in $o_i(t)$ as 

\begin{equation}
d_{k, l} = |t_{k}^{(r)} - t_{l}^{(o)}| 
\end{equation}

where $t_{k}^{(r)}$ is the time of the $k$th spike in the reference spike train and $t_{l}^{(o)}$ represents the time of the $l$th spike in the observed spike train,

The cost matrix $\Lambda$ is determined with the following recursion relation
\begin{equation}
    \Lambda_{k,l} = \min \{ \Lambda_{k-1,l-1} + d_{k,l}, \Lambda_{k-1,l} + a_{cap}, \Lambda_{k,l-1} + a_{cap}, \Lambda_{k-1,l-1} + 2 a_{cap} \}
\end{equation} and base cases 
\begin{equation}
\Lambda_{0,l} = l \ a_{cap}
\end{equation}
\begin{equation}
\Lambda_{k,0} = k \ a_{cap}
\end{equation}
where the parameter $a_{cap}$ represents the maximum temporal separation between two spikes that can be considered paired. We chose to set $a_{cap}$ at 15 timesteps, which corresponded to roughly 45 ms. 
If the first argument of the right hand side in equation (4) is minimum, then spike $k$ and spike $l$ should be paired to minimize the cost of $\Lambda_{k,l}$. The second argument in equation (4) is minimum if it is cheapest to leave spike $k$ in $r_i(t)$ unpaired, incurring an additional cost of $a_{cap}$. Likewise, the third argument in (4) is minimal if it is cheapest to leave spike $l$ in $o_i(t)$ unpaired. Lastly, if it is cheapest to leave both $k$ and $l$ without pairs then we incur the additional cost of $2 a_{cap}$. 

Once the cost matrix $\Lambda$ is completely determined, we backtrack through the matrix from position $[n,m]$ to one of the base case positions: $[k,0]$ for some $k$ or $[0,l]$ for some integer $l$. As we backtrack through positions in the matrix, we move in the direction  along which $\Lambda_{n,m}$ was defined with the recurrence relation $(4)$. This corresponds to identifying the optimal set of spike pairs $P$ where $[k,l] \in P$ if and only if the $k$th spike of $r_i(t)$ is paired with the $l$th spike of $o_i(t)$ to minimize the total cost.

Equipped with the set of spike pairs $P$ that minimize total cost, we can then determine which spikes in $r_i(t)$ do not have a counterpart in $o_i(t)$ and vice-versa. For every unpaired spike in $r_i(t)$, we attempt to induce a spike at $o_i(t)$ by increasing some of the pre-synaptic weights of neuron $i$ in the optimized network (Figure \ref{PSPM_algorithm_figure}). Let $t_{k}^{(R)}$ be the time of the unpaired spike in $r_i(t)$. To choose which inbound synaptic connections $W_{ij}$ should be increased to induce a spike in $o_i(t)$ are determined by which neurons $j$ in the presynaptic pool spiked within the interval $[t_{k}^{(R)} - z, t_{k}^{(R)}]$ where $z$ is an integer number of timesteps. For each of these neurons $j$ we make the following update 
\begin{equation}
    W_{i j} \leftarrow W_{i j } + \delta
\end{equation}
where $\delta$ is drawn from $[0, 10^{-7}]$ V and $z$ was set to 10 timesteps. 

In the case of an extra spike in $o_i(t)$, a similar procedure is followed, save that relevant weights $W_{ij}$ are stochastically decreased. Note that weight matrix values are not allowed to change sign, which would correspond to an excitatory synapse becoming inhibitory or vice-versa. Instead, if an excitatory weight is decreased below zero or an inhibitory weight is increased above zero, the weight is simply set to zero.\ 

In addition, homeostatic weight modifications are made to keep the entire network at a desired level of activity. In the event of excess spiking throughout the observed spike trains $\mathcal{O}$, all weights (not just the excitatory weights) of the observed network are stochastically diminished by subtracting a small random number from each of the weights. If the observed network produces an inadequate number of spikes, all weights are stochastically increased by adding small random numbers to each of the weights. Further, changes to weights are additive increases and decreases, rather than multiplicative increases or decreases in the magnitude of the weight, though we do not allow inhibitory connections to become excitatory, nor vice versa. If the reference spike trains $\mathcal{R}$ contain $x$ spikes and the observed spike trains $\mathcal{O}$ contain $y$ spikes, changes to each weight value $W_{ij}$ are determined by drawing from a uniform distribution over $[0 , (x-y) \times 10^{-11}]$ V.\\ 

\subsection*{Spike Train Similarity Measures}
To assess agreement between spike trains after the algorithm was run, we used a modification of the van Rossum distance metric \cite{vanrossum}. In the original van Rossum paper, spike trains are filtered with an exponential window to smooth the signal. To generate a distance measure between the spike trains, absolute or squared differences between convolved signals are then integrated over the duration of the signals. In this project it was desirable to have tolerance on both sides of a spike (before and after), so instead of using an exponential windowing function, we convolved binary spike trains with a Gaussian window as described in Schreiber et. al \cite{Schreiber}. Our Gaussian filter had a mean $\mu = 0$ and a standard deviation of $\sigma = 5\sqrt{2}$ time-steps or roughly $\sigma \approx 21$ ms. The kernel, or windowing function, $K(t) = e^{- \frac{1}{2\sigma^2} t^2}$ is convolved with the $i$th binary spike train $s_i(t) \in \{0,1\}^T$, which is defined for the time interval of the simulation $[0,T]$ where $T$ is the total number of time steps in a simulation. The result of this convolution is essentially a sum of Gaussians and is defined as the \textit{activity signal for neuron $i$}  
\begin{center}
$a(s_i(t),t) = K(t) * s_{i}(t) = \int\limits_{0}^{t} d\tau \ e^{-(t-\tau)^2/2{\sigma}^2} s_i(\tau)$. \end{center} 
A total network activity signal can be found by summing individual activity signals $a(s_i(t),t)$ over each neuron index $i$ with $N$ the total number of neurons. If $\mathcal{S} = [s_1(t), s_2(t)...,s_i(t),...,s_N(t)] \in \{0,1\}^{T \times N}$ is the matrix containing each of the output spike trains of a network, then the \textit{total activity signal for spike trains $\mathcal{S}$} is
\begin{center}
$A(\mathcal{S},t) = \sum\limits_{i=1}^{N} \ a(s_i(t),t)$
\end{center}
We will adopt the convention that lowercase $a(s_i(t),t)$ corresponds to the activity signal of neuron $i$ while the upper case $A(\mathcal{S}, t)$ corresponds to the total activity signal of the network spike trains $\mathcal{S}$.

From these two types of activity signals, two distance metrics are established. First a pairwise distance metric $D_P(\mathcal{S},\mathcal{R})$ is constructed by comparing the individual activity signals of two sets of spike trains (with the same number of neurons $N$) $\mathcal{S},\mathcal{R} \in \{0,1\}^{T \times N}$, \textit{neuron by neuron} over the time interval $[0,T]$.
If $s_i(t)$ and $r_i(t)$ are the $i$th binary spike trains for distinct network outputs $\mathcal{S}$ and $\mathcal{R}$, then the \textit{pairwise distance measure} would be 
\begin{center}
$D_P(\mathcal{S},\mathcal{R}) = \sum\limits_{i=1}^{N} \ \int\limits_{0}^{T} dt  \ [a(s_i(t),t) - a(r_i(t), t)]^2 $. 
\end{center}
We also define an \textit{aggregate distance measure} between the two sets of spike trains $\mathcal{S}$ and $\mathcal{R}$ as 
\begin{center}
$D_A(\mathcal{S},\mathcal{R}) = \int\limits_{0}^{T} dt \ [A(\mathcal{S},t) - A(\mathcal{R},t)]^2 $. 
\end{center}
\noindent
where $A(\mathcal{S},t)$ and $A(\mathcal{R},t)$ are total activity signals of spike trains $\mathcal{S}$ and $\mathcal{R}$ respectively. 
\indent

In addition to these spike train distance measures, we also assessed agreement between spike trains by comparing inter-spike-interval (ISI) distributions, which contain information about the regularity of spiking in a network. For a given spike train $s_i(t)$, the inter-spike interval distribution is a series of observations of the number of time steps between adjacent spikes. For instance, if at time $t$, $s_i(t)=1$ and the time of the next spike is $t+a$ so $s_i(t+a)=1$ and $s_i(t+b)=0$ for $0<b<a$, then the value $a$ is appended to the ISI distribution. For a network wide measure ISI distribution, observations from each neuron are concatenated to produce a distribution with the entire set of observations.

\begin{figure}
	\centering
		\includegraphics[width=0.9\textwidth]{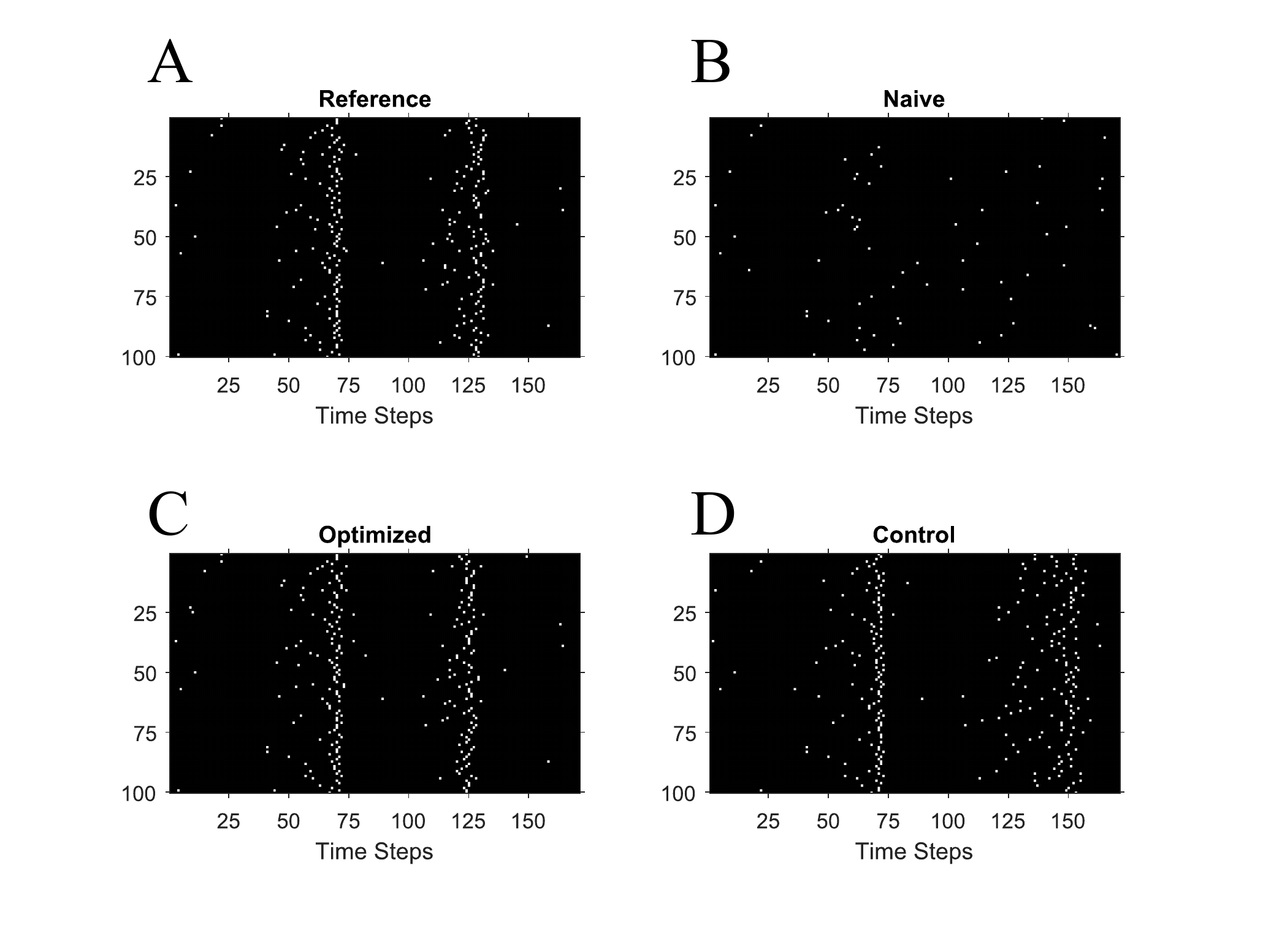}
	\caption{PSPM improves qualitative agreement in output spike trains. (\textbf{A}) The outputs $\mathcal{R}$ the reference network for a representative trial when stimulated with Gaussian input currents. Neuron identity is plotted on the y-axis, while time is plotted on the x-axis. These spike trains serve as the goal output firing activity throughout the application of the algorithm. (\textbf{B}) The outputs $\mathcal{N}$ of a naïve network simulated with identical inputs as the reference network used in (\textbf{A}). Note that the dissimilarity in outputs between (\textbf{A}) and (\textbf{B}) is due entirely to differences in the weight matrices of the reference and naive networks. (\textbf{C}) After optimization of the weight matrix, the network’s firing activity $\mathcal{O}$ is qualitatively similar to that of the reference network. (\textbf{D}) The spiking $\mathcal{C}$ of the control network also differs from the reference. The weights of the control network are changed each time the algorithm would have made a change to a specific weight in the optimized network, but at a different random synapse instead. Thus, the disparity in performance between the optimized and control networks is due to PSPM's neuron-by-neuron precise weight adjustments.}
	\label{spike_trains}
\end{figure}

\begin{figure}
	\centering
		\includegraphics[width=.8 \textwidth]{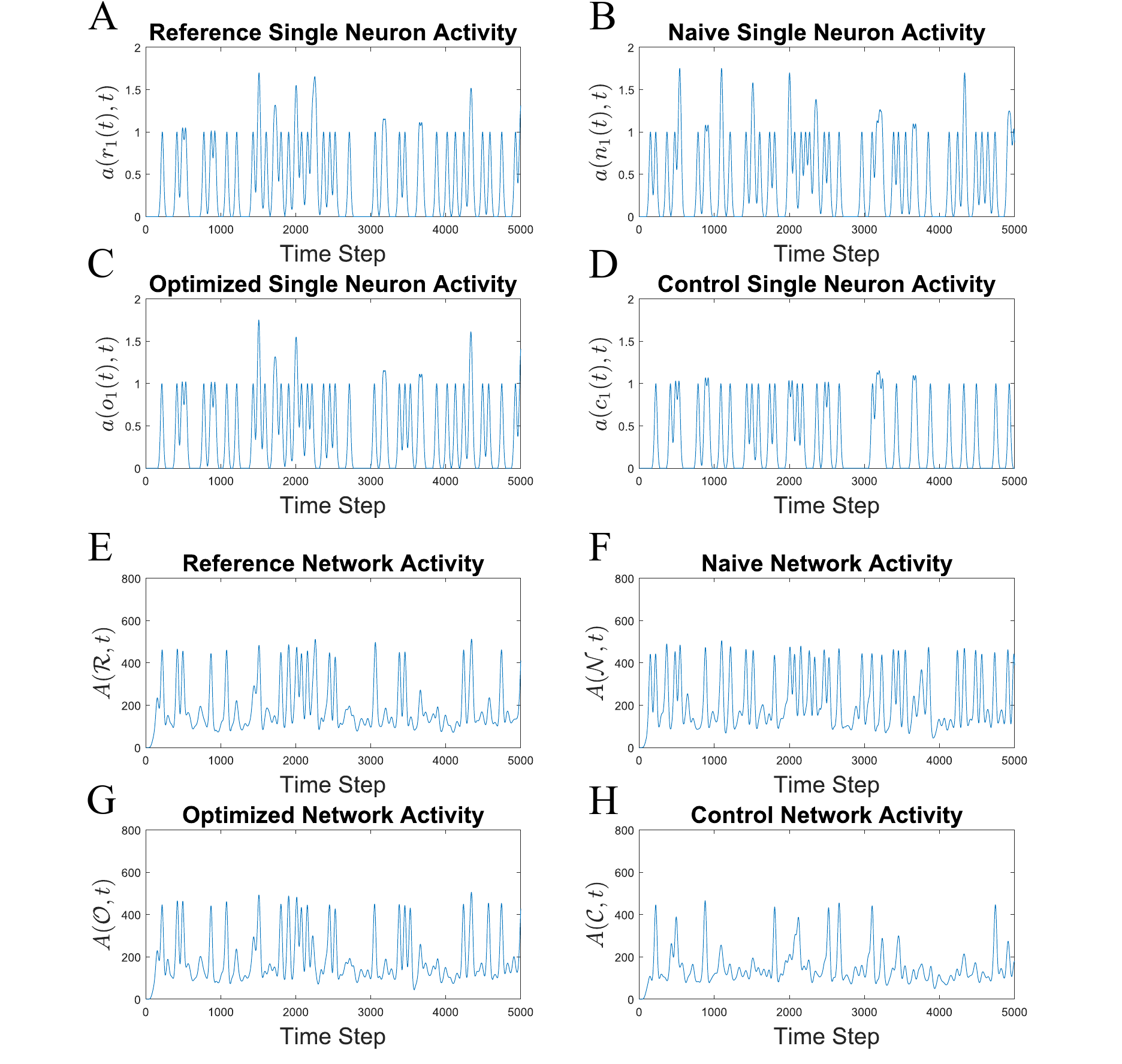}
        
	\caption{PSPM improves qualitative agreement between output activity signals and reference activity signals (Methods). Subplots (\textbf{A})-(\textbf{D}) show single neuron activity signals of network outputs for a representative trial. In particular, these are the activity signals for neuron 1 in this trial. If $r_1(t)$, $n_1(t)$, $o_1(t)$, $c_1(t)$ represent the spike trains of the first neuron in the reference, naive, optimized, and control networks respectively then the activity signals $a(r_1(t),t)$, $a(n_1(t),t)$, $a(o_1(t),t)$, $a(c_1(t),t)$ are plotted in (\textbf{A}), (\textbf{B}), (\textbf{C}), and (\textbf{D}) respectively (Methods). Transitioning from the naive state to the optimized state there is clear improvement in the agreement with the reference activity signal. Similarly, the post-optimization signal $a(o_1(t),t)$ exhibits greater agreement with the reference signal $a(r_1(t),t)$ than does the control total activity signal $A(\mathcal{C},t)$ with the reference signal. The fact that (\textbf{A}) and (\textbf{C}) show closest agreement indicates successful learning due to PSPM. Total activity signals are plotted in (\textbf{E})-(\textbf{H}). If the total activity signals for the reference spike trains $\mathcal{R}$, naive spike trains ${\mathcal{N}}$, optimized spike trains ${\mathcal{O}}$, and control spike trains $\mathcal{C}$ are represented by $A(\mathcal{R},t)$, $A(\mathcal{N},t)$, $A(\mathcal{O},t)$, $A(\mathcal{C},t)$, then these are respectively plotted under labels (\textbf{E}), (\textbf{F}), (\textbf{C}), and (\textbf{D}) (Methods).}
	\label{activity_signal}
	
\end{figure}

\begin{figure}
	\centering
		\includegraphics[width=\textwidth]{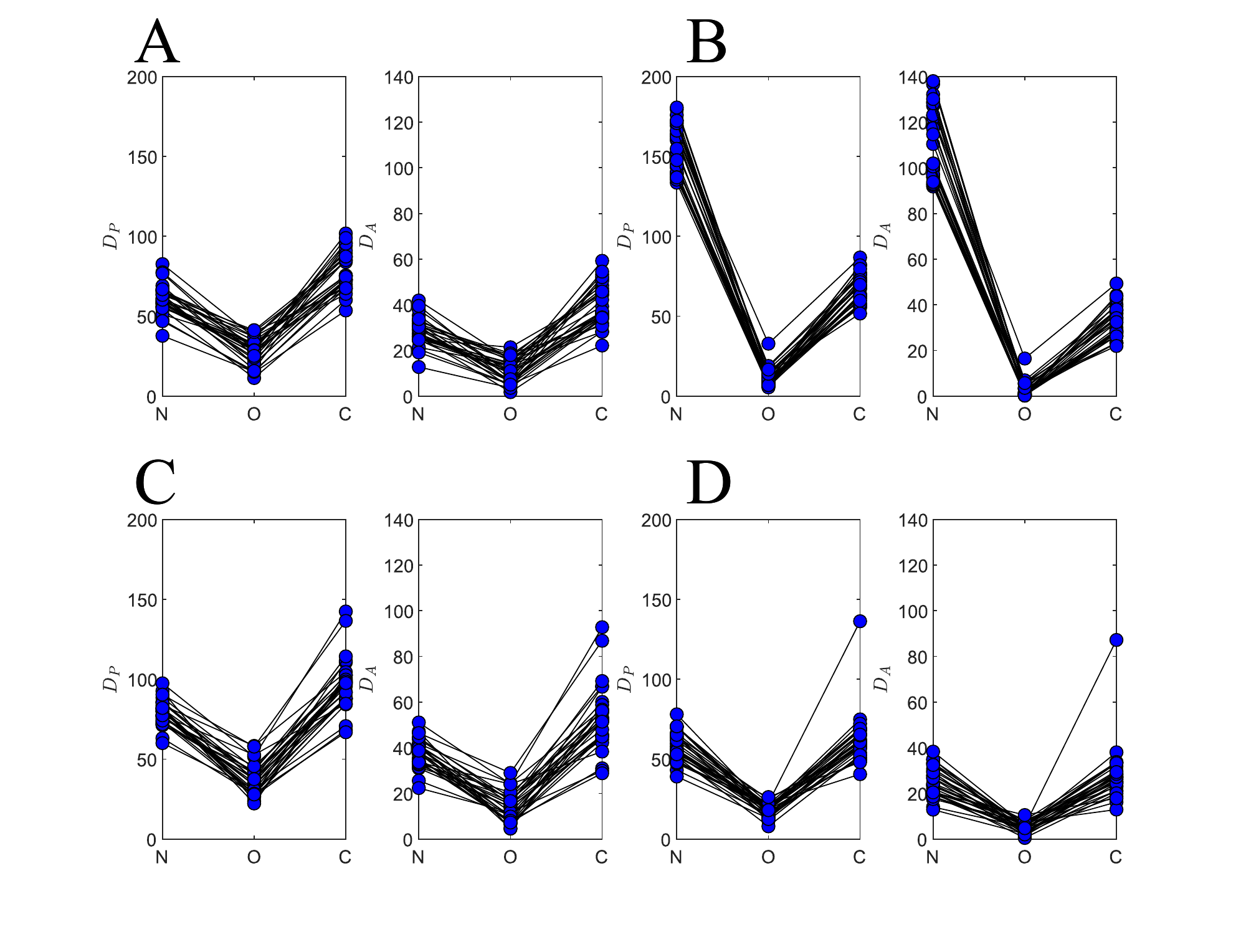}
	\caption{PSPM decreases distance measures for all initial network configurations. (\textbf{A}) Pairwise $D_P(\cdot, \mathcal{R})$ and aggregate $D_A(\cdot, \mathcal{R})$ distance measures from 30 trials for where initial weight matrix magnitudes were drawn from a \textit{uniform} distribution on (Methods). Moving from left to right in each panel, distances are shown for the naive $\mathcal{N}$ (left), optimized $\mathcal{O}$ (middle), and control $\mathcal{C}$ network outputs (right).  (\textbf{B}) Distance measures for a reference weight distribution with magnitudes drawn from a \textit{gaussian} distribution (see Methods). (\textbf{C}) Distance measures for \textit{sparse} initial configuration. (\textbf{D}) Distance measures for reference outputs produced from a \textit{naive-half-max} initial configuration (Methods). Note that in each of these figures the optimized distances are consistently smaller than those of the naive or control, indicating successful learning due to PSPM prescribed weight changes.}
	\label{distance_measure}
\end{figure}

To compare ISI distributions between two output spike trains, we conduct a two-sample Kolmogorov-Smirnoff test. This non-parametric statistical test measures the probability that two sets of observations are drawn from the same distribution by evaluating the largest disparity in the respective cumulative probability distributions. A small p-value warrants rejection of the null hypothesis which is that the two observations are drawn from the same distribution. Thus a high p-value indicates a high goodness of fit between two ISI distributions.

\begin{figure}
	\centering
    	\includegraphics[width=\textwidth]{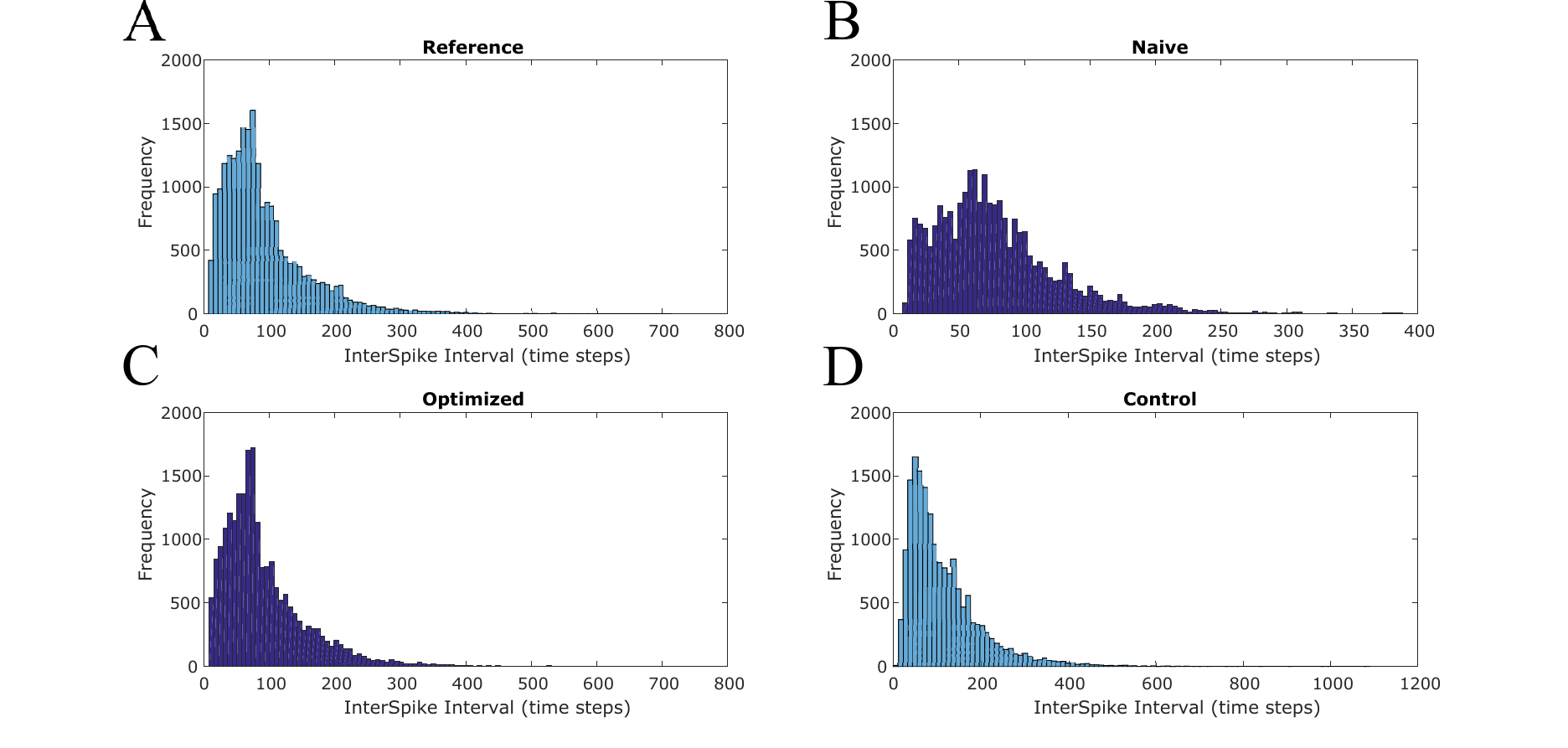}
	\caption{PSPM Learning improves agreement of inter-spike interval (ISI) distributions. (\textbf{A}) The network-wide ISI distribution for the reference network most closely resembles the ISI of the optimized network (\textbf{C}) for this trial. The naive (\textbf{B}) and control (\textbf{D}) network ISI distributions both differ qualitatively from the reference ISI distribution. Thus the synapse specific changes called for by PSPM produces an ISI distribution similar to the reference.}
	\label{ISI_dist}
\end{figure}

\begin{figure}
	\centering
		\includegraphics[width=\textwidth]{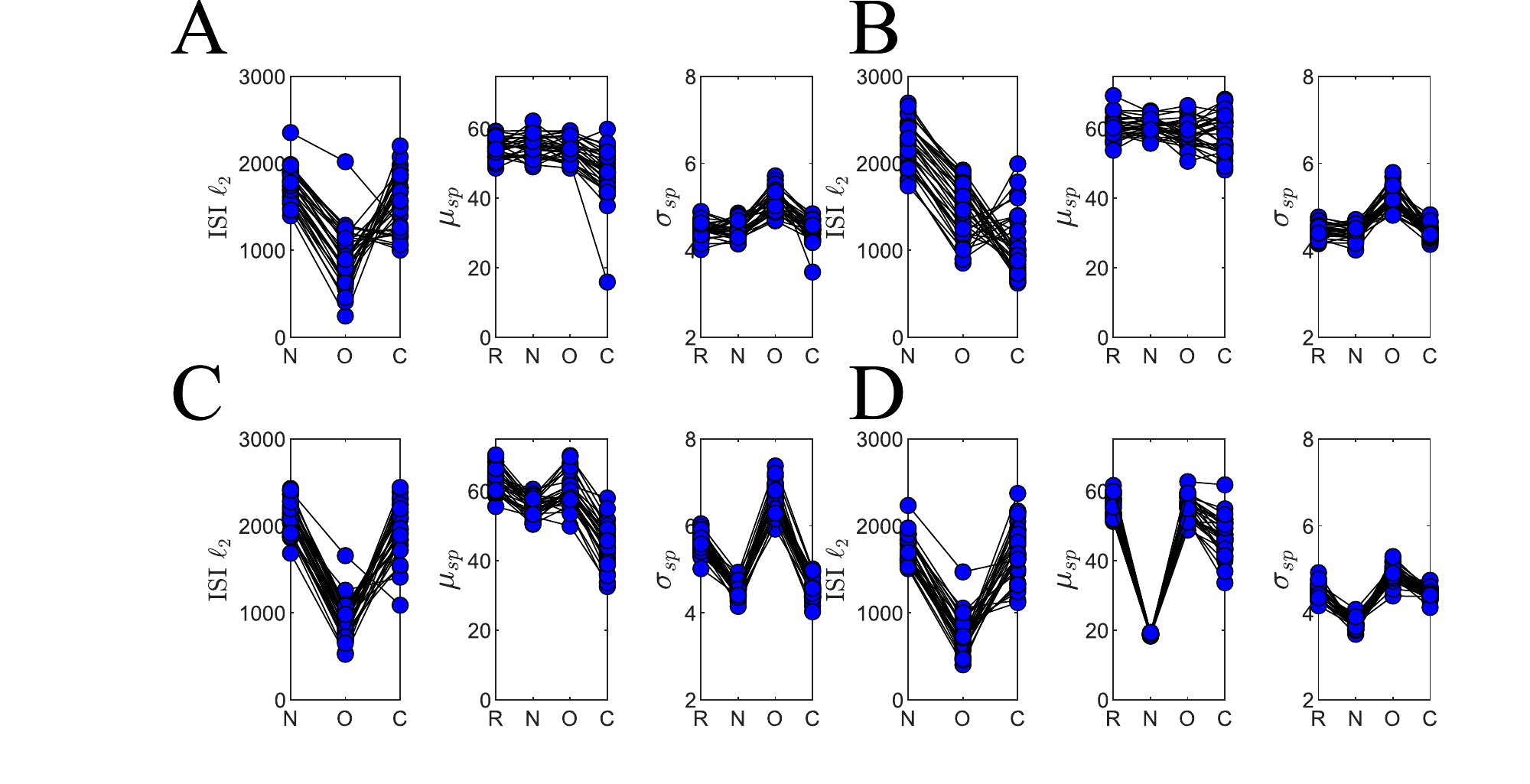}
	\caption{Improvement in ISI goodness of fit and basic spike statistics dependent on initial network configuration. (\textbf{A}) $\ell_2$ distance measures between ISI histograms and basic spike statistics for \textit{uniform} initial reference and naive networks (Methods). The plots are organized left to right for reference, naive, optimized, and finally control C conditions. The ISI goodness of fit improves during optimization indicating similarity in the reference and optimized inter-spike interval distributions. For the \textit{uniform} initial distributions, the control ISI $\ell_2$ distances are than those of the optimized networks, but for all other initializations, the ISI $\ell_2$ distance is lowest for the optimized contdition. Means and variances of spike number are computed by recording the number of spikes in each neuron's spike train and taking averages and variances over neurons. While the mean spike number is roughly unchanged through optimization (from naive to optimized), the variance in spike number diverges from the reference value. (\textbf{B}) For the \textit{gaussian} initial network configuration, ISI goodness of fit does not improve with optimization nor do the basic spike statistics. (\textbf{C}) For the \textit{sparse} initial network configuration, PSPM improves network performance for both ISI goodness of fit and mean spike number but not for variance. (\textbf{D}) The \textit{naive-half-max} initial configuration enjoys benefits in ISI goodness of fit and mean spike number but shows spike number variance again increasing with application of PSPM. }
	\label{spike_statistics}
\end{figure}

\subsection*{Probabalistic Integrate-and-Fire (PIF) Model}

In our second set of simulations, reference binary spike trains $\mathcal{R}$ were generated from a probabalistic integrate and fire (PIF) model with $N=400$ neurons \cite{karimipanah2}-\cite{Larremore2}. Like the synapses in the LIF, the strength of PIF synapses are represented by a weight matrix $W \in \mathbb{R}^{N \times N}$. To impose sparsity, we set the probability that a synapse is created between any two neurons $i$ and $j$ to $p=10\%$. In our PIF network, all synapses were excitatory. The weight values were drawn from a uniform distribution on $[0,0.02]$ so that the mean (nonzero) weight value was $0.01$. The maximum eigenvalue $\lambda$ was calculated and each element of the weight matrix $W$ was subsequently divided by $\lambda$ so that the maximum eigenvalue of $W$ is 1 and the network operates at criticality \cite{karimipanah2}. Input values $I_i(t)$ were drawn from a Poisson distribution with mean and variance $\mu=\sigma^2=1$ which were then multiplied by 0.001 to produce a network wide average spiking frequency greater than $0.003$ spikes per time-step or $1$ Hz if each time-step is 3 ms, consistent with the overall spike rate of our LIF model.  The binary state of neuron $i$ at time $t+1$, $s_i(t+1)$, is given in terms of the state of all other neurons at time $t$:  

\begin{equation}
s_i(t+1) = \Theta[\sum \limits_{j=0}^{N} W_{ij} s_j(t) + I_i(t) - \xi_i(t)]
\end{equation}

where $\xi_i(t)$ are random numbers drawn from a uniform distribution on $[0,1]$ and $\Theta$ is the unit step function. Whereas the LIF network parameters have dimensions corresponding to voltage and current, the PIF model parameters are dimensionless. This PIF network was simulated for $T=50,000$ time-steps to produce the reference spike trains $\mathcal{R}=[r_1(t),...,r_i(t),...,r_N(t)] \in \{0,1\}^{T \times N}$. This simulation time was chosen so that there were 5000 or more avalanches (\textit{Avalanche Analysis}). The PIF outputs $\mathcal{R}$ were then broken up into 5 spike trains of 10,000 timesteps each of which served as a set of reference spike trains for the PSPM learning procedure. These spike trains were split so as to reduce the algorithm's input size, improving the speed at which a solution could be obtained. Naive LIF networks were initialized with weight magnitudes drawn from a uniform distribution on $[0,1 \times 10^{-3}]$ with all synapses designated as excitatory.\\

\subsection*{Avalanche Analysis}
To test whether PSPM successfully induced criticality in the optimized LIF networks, avalanche statistics were calculated for each of the output spike trains $\mathcal{R}$, $\mathcal{N}$, $\mathcal{O}$, and $\mathcal{C}$. From a set of spike trains $s_i(t)$ for neurons $i$, a summed network spiking $F(t) = \sum \limits_{i} s_i(t)$ was evaluated. Avalanches were defined as events where the summed network spiking $F(t)$ exceeded the $20$th percentile of the all summed activity values over the simulation interval $[0,T]$. An avalanche persists from the time step $F(t)$ first passes above the $20$th percentile threshold until $F(t)$ sinks below this threshold. This percentile threshold was chosen so that in a simulation of $T=50,000$ time steps, the number of avalanches exceeded 5000 for the PIF network outputs $\mathcal{R}$.  For each avalanche, the size $S$ and duration $D$ were recorded. Size is defined as the number of spikes within an avalanche while duration is the number of time steps for which the avalanche persists. As described in \cite{karimipanah1}, a maximum likelihood method was employed to fit power law probability distributions for the sizes and durations of avalanches. Namely that the size distribution follows $P(S) \sim  S^{-\tau}$ while the duration distribution follows $P(D) \sim D^{-\alpha}$. At criticality, the average avalanche size $<S>$ and duration $D$ of avalanches also obeys a power law $<S> \sim D^{-\beta}$ with critical exponent $\beta$. By using the calculated critical exponents $\alpha$ and $\tau$ from the size and duration probability densities one can calculate a predicted $\beta_p = (\alpha - 1)/(\tau - 1)$ which can then be compared to an empirically observed critical exponent $\beta_o$ obtained by fitting average size $<S>$ and duration $D$ observations for the avalanches in the output spike train.

\section*{Results}
\subsection*{PSPM Learning Improves Spike Train Distance Measures}
In simulations with LIF reference and naive networks generated from the distributions described in \textit{Initial Network Parameters}, optimization improved agreement between $\mathcal{O}$ and $\mathcal{R}$. Figure \ref{spike_trains} (\textbf{A})-(\textbf{D}) show sample raster plots of $\mathcal{R}$, $\mathcal{N}$, $\mathcal{O}$, and $\mathcal{C}$ for a representative trial with \textit{uniform} initial network parameters (Methods).  $\mathcal{R}$ and $\mathcal{N}$ are similar in the overall level of network activity, but close inspection reveals significant differences in the timing of spikes for individual neurons in the network. Because both networks were stimulated with the same input currents, the difference in the reference and naive weight matrices account for the observed disparity in the output spike trains. $\mathcal{O}$, however, exhibits considerably better qualitative agreement with $\mathcal{R}$ than $\mathcal{N}$ does, indicating improvement due to PSPM learning. $\mathcal{C}$, despite benefiting from similar homeostatic adjustments to its weight matrix as the optimized weight matrix (Methods), shows disagreement with $\mathcal{R}$ due to the weight changes made at random synapses during learning. This indicates that network performance is sensitive to the prescribed local weight changes called for by our PSPM learning rules. 

\indent
To corroborate these qualitative assessments of spike train similarity for this trial, we generate network activity signals with the Van-Rossum like method described in \textit{Spike Train Similarity Measures}. Figure \ref{activity_signal} (\textbf{A})-(\textbf{D}) shows the single neuron activity signals for an example neuron, in this case, the first neuron in the network. Again, the optimized activity signal $a(o_1(t),t)$ shows better agreement with the reference $a(r_1(t),t)$ than either the naive $a(n_1(t),t)$ or control $a(c_1(t), t)$ activity signals do. In addition to single neuron signals, we show the total network activity signals for this trial in Figure \ref{activity_signal} (\textbf{E})-(\textbf{H}) to visually capture network-wide spike train behavior. As before, the optimized network activity $A(\mathcal{O},t)$ shows the greatest agreement with the total activity signal of the reference $A(\mathcal{R},t)$. 


Distance measure results for each of the initial network configurations described in \textit{Initial Network Configurations} are shown in Figure \ref{distance_measure}. For each initial configuration, 30 trials were conducted. Tables 1 and 2 show these pairwise and aggregate distance measures respectively. For each of the initial network configurations, the distance measures of the optimized networks are lower than the distance measures for the naive and control networks. The fact that PSPM improves performance of the optimized network but not the control demonstrates the importance of making local weight updates at appropriate synapses during learning (Methods).

\begin{table}
\begin{center}
    \begin{tabular}{ | l | l | l | l | l | l | p{5cm} |}
    \hline
    Initial Configuration & $D_P(\mathcal{N},\mathcal{R})$ & $D_P(\mathcal{O},\mathcal{R})$ & $D_P(\mathcal{C},\mathcal{R})$ \\ \hline
    \textit{uniform} & 
    $58 \pm 9$ & 
    $18 \pm 4$ & 
    $63 \pm 16$ 
  \\ \hline
    \textit{gaussian} & 
    $61 \pm 9$ & 
    $28 \pm 8$ & 
    $79 \pm 12$,  \\ \hline
    \textit{sparse} & 
    $79  \pm 8$ & 
    $37 \pm 0.425$ & 
    $97 \pm 16$  \\
    \hline
    \textit{naive-half-max} & $154 \pm 15$ & 
    $11 \pm 5$ & 
    $68 \pm 9$ 
    \\
    \hline
   
    \end{tabular}

\caption{Pairwise distance measures for each of the 4 initial network configurations. Mean and standard deviation reported for 30 trials. }

\end{center}

\begin{center}
    \begin{tabular}{ | l | l | l | l | l | l | p{5cm} |}
    \hline
    Initial Configuration & $D_A(\mathcal{N},\mathcal{R})$ & $D_A(\mathcal{O},\mathcal{R})$ & $D_A(\mathcal{C},\mathcal{R})$ \\ \hline
    \textit{uniform} &
    $25 \pm 7$ & 
    $5 \pm 2$ & 
    $28 \pm 13$ 
  \\ \hline
    \textit{gaussian} & 
    $28 \pm 6$ & 
    $11 \pm 5$ & 
    $40 \pm 9$,  \\ \hline
    \textit{sparse} & 
    $37 \pm 6$ & 
    $14 \pm 6$ & 
    $53 \pm 14$  \\
    \hline
    \textit{naive-half-max} & $112 \pm 15$ & $3 \pm 3$ & $33 \pm 7$ 
    \\
    \hline
  
    \end{tabular}
    
\caption{Aggregate distance measures for each of the 4 initial network configurations. Mean and standard deviation reported for 30 trials.}    
    
\end{center}
\end{table}

\subsection*{Improved agreement in ISI distributions and basic spike statistics dependent on Initial Network Configuration}

In addition to improving spike train similarity, the PSPM procedure also improves goodness of fit between inter-spike interval (ISI) distributions. For example, the ISI distributions for each of the four spike trains $\mathcal{R}$, $\mathcal{N}$, $\mathcal{O}$, and $\mathcal{C}$ for a sample trial are shown in Figure \ref{ISI_dist} (\textbf{A})-(\textbf{D}). Qualitatively, the ISI distribution of the optimized network shows best agreement with that of the reference network. The ISI distributions of the naive and control networks differ somewhat from that of the reference network. \\
To quantify similarity between ISI distributions, two-sample Kolmogorov-Smirnov (KS) tests were conducted between the set of reference ISI observations and the observed ISI values from the naive, optimized, and control outputs (Methods). Figure \ref{spike_statistics} shows the KS test p-values recorded for each of the 30 trials along with spike number mean and variance (computed over neurons in the network). The ISI distribution for $\mathcal{O}$ demonstrates the closest fit with that of $\mathcal{R}$ as indicated by the high mean p-value for each of the initial network configurations.

\begin{table}[]
    \centering
    \begin{tabular}{|c|c|c|c|}
        \hline
        distribution & Naive $\ell_2$ & Optimized $\ell_2$ & Control $\ell_2$  \\
        \hline
        \textit{uniform} &  $17 \pm 2$ &
         $9 \pm 10$ & 
         $16 \pm 20$  \\
        \hline
        \textit{gaussian}  & $22 \pm 2$ & 
        $15 \pm 10$ & 
        $11 \pm 10$  \\
        \hline
        \textit{sparse}  & $21 \pm 2$ & 
        $8 \pm 10$ & 
        $20 \pm 20$  \\
        \hline 
        \textit{naive-half-max}  & 
        $17 \pm 20$ & 
        $7 \pm 10$ & 
        $17 \pm 20$  \\
        \hline
    \end{tabular}
    \caption{ISI goodness of fit. Actual Values are $10^2$ times those reported above. }
    \label{ISI_table}
\end{table}

KS test p-values for each distribution are shown in Table \ref{ISI_table}. The mean $p_{KS}$ values are highest for the ISI distribution of the optimized network demonstrating best average agreement between the reference ISI distribution and that of the optimized network.  This indicates that the PSPM algorithm improves agreement between ISI distributions of the reference and optimized network. However, the large standard deviations in $p_{KS}$ for each network and initial configuration indicates a large trial to trial variability. Learning the ISI distribution appears sensitive to the initial weights of the naive and reference networks in each trial.\\
\indent
In addition, spike number mean and variance show sensitivity for the initial network conditions (Figure \ref{spike_statistics}). Mean spike numbers were calculated for a collection of spike trains $\mathcal{S} = [s_1(t), s_2(t)...,s_i(t),...,s_N(t)] \in \{0,1\}^{T \times N}$ by first counting the number of spikes in each individual spike train $s_i(t)$ and then taking an average over neurons $i$. Variance was similarly calculated over the spike numbers in each individual neuron's spike train. While the mean spike number varies sporadically throughout optimization, spike number variance tends to increase with application of PSPM. Interestingly, the \textit{gaussian} and \textit{sparse} networks show a larger spike number variance for the optimized outputs $\mathcal{O}$ than for the control outputs $\mathcal{C}$, indicating that for these distributions, network-wide homeostatic adjustments do not considerably influence the spike number variance but the targeted spike-matching weight adjustments due to PSPM increase spike number variance. \\ 

\subsection*{PSPM does not induce agreement between weight matrices}

With the PSPM algorithm's demonstrated success in reducing spike train distance measures and improving goodness of fit of ISI distributions, a plausible expectation may be that the algorithm improves similarity between the reference and optimized weight matrices $W^{(R)}, W^{(O)} \in \mathbb{R}^{N \times N}$ but an analysis of the resulting weight matrices shows that this is not the case. To calculate the similarity between weight matrices, component-wise sums of squares, also known as Frobenius norms, were calculated for each of the following matrices $W^{(N)} - W^{(R)}$, $W^{(O)}-W^{(R)}$, $W^{(C)} - W^{(R)}$. These errors are shown in Table 3 averaged over 30 trials for each of the initial network configurations. The optimized weights $W^{(O)}$ showed the largest disagreement with the reference weights $W^{(R)}$ for the \textit{uniform}, \textit{gaussian}, and \textit{sparse} and only shows slight improvement for the \textit{naive-half-max} condition where the elements of the naive and reference weight matrices, $W^{(N)}$ and $W^{(R)}$ respectively, are drawn from distinct probability distributions. A possible explanation for this disagreement between optimized and reference weight matrices despite improvement in distance measures and ISI statistics is that many possible weight matrices could produce the same spike trains. 


\begin{table}
\begin{center}
    \begin{tabular}{ | l | l | l | l | l | l | p{5cm} |}
    \hline
    Initial Configuration & $\sum\limits_{ij}{(W_{ij}^{(N)} - W_{ij}^{(R)})^2}$ & $\sum\limits_{ij}{(W_{ij}^{(O)} - R_{ij}^{(R)})^2}$ & $\sum\limits_{ij}{(W_{ij}^{(C)} - W_{ij}^{(R)})^2}$ \\ \hline
    \textit{uniform} & $.0643 \pm .0224$ & $.0955 \pm .1524$ & $.0618 \pm .0051$ 
  \\ \hline
    \textit{gaussian} & $.0326 \pm .0001$ & $.0559 \pm .0027$ & $.0434 \pm .0009$,  \\ \hline
    \textit{sparse} & $.1799 \pm .006$ & $.1916 \pm .0034$ & $.1813 \pm .0007$  \\
    \hline
    \textit{naive-half-max} & $.0599 \pm .0003$ & $.0593 \pm .0017$ & $.0506 \pm .0002$ 
    \\
    \hline
  
    \end{tabular}
\end{center}
\caption{Weight Matrix Differences for each initial network configuration. Mean and standard deviation reported for 30 trials.}
\end{table}

\subsection*{PSPM used to generate critical behavior in a LIF network}

In the second set of simulations, a probabilistic integrate-and-fire (PIF) networks was generated with the maximum eigenvalue of the weight matrix tuned to 1 (Methods). The PIF network was driven with Poisson inputs and the outputs were evaluated with avalanche analysis. LIF networks were then optimized to reproduce the critical outputs of the PIF, but in a deterministic model network. The spike trains of this PIF network were simulated and subsequently used as the reference data for runs of our algorithm. Namely, the 50,000 time steps of the PIF simulation were split into five sets of 10,000 time steps, each of which was used for a run of the PSPM algorithm. The resulting output spike trains were concatenated to provide adequate data for critical avalanche analysis. While the PIF network received external, scaled Poisson inputs (Methods), the LIF network was stimulated using a Gaussian input current distribution identical to that used in our first set of simulations. Poisson inputs were only used for the PIF network on account of its different dynamics. Input currents in the PIF represent probabilities of firing due to external input, whereas input currents in the LIF change the membrane potential. Unlike the first set of simulations, the LIF network synapses were exclusively excitatory, as was the case for the PIF networks described by Karimipanah \cite{karimipanah2}\\

\begin{figure}
	\centering
		\includegraphics[width=\textwidth]{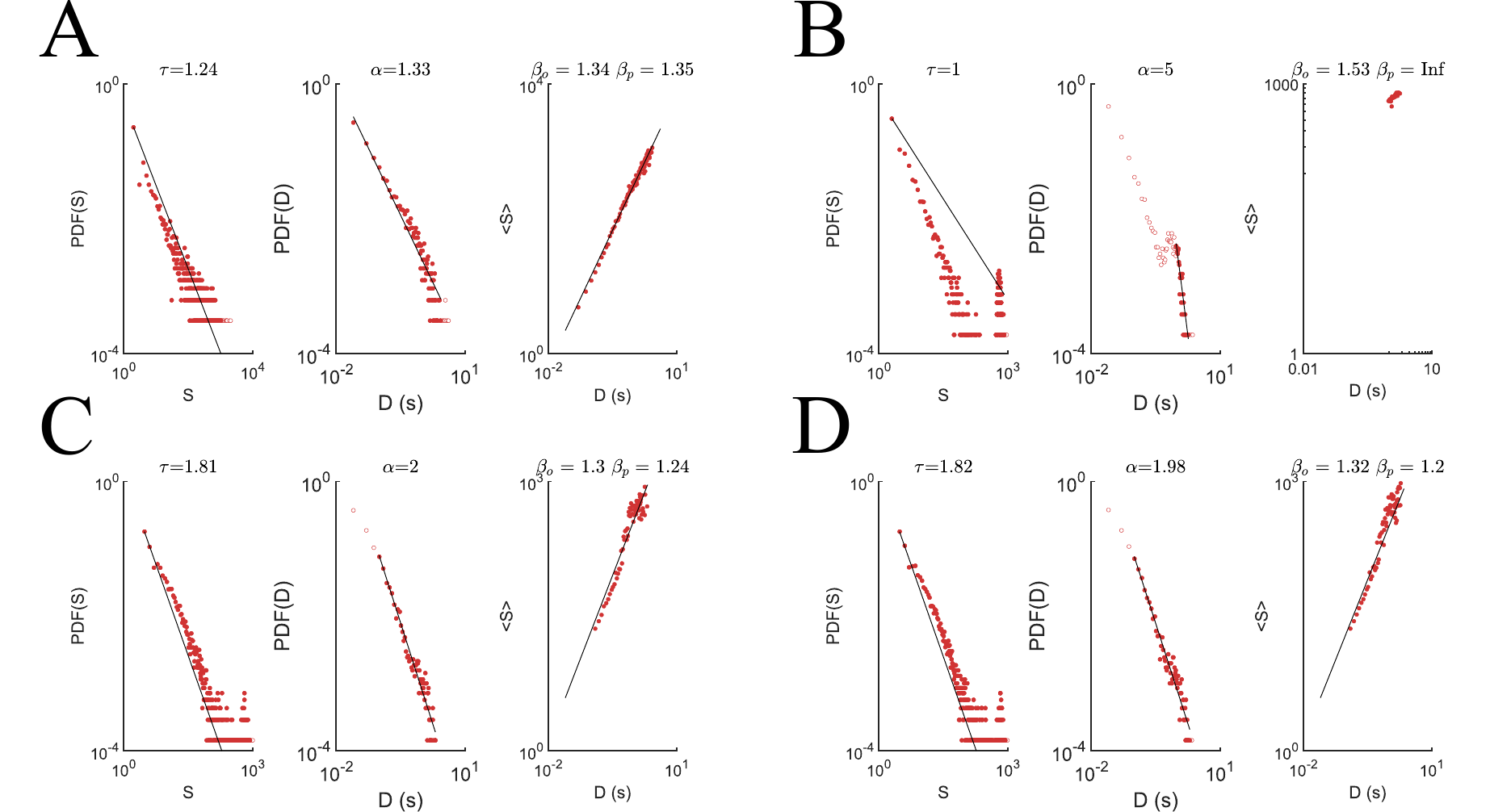}
	\caption{PSPM used to generate a LIF network operating near criticality. (\textbf{A}) Scaling relations for the avalanches of the PIF outputs. The size (left) and duration (middle) distributions of the avalanches present in the PIF outputs (Methods). Data is presented in log-log plots so that the empirically fit power laws are clear. The critical exponents for the size $\tau = 1.242$ and duration $\alpha=1.327$ are fit through regression and used to predict a theoretical value for the scaling constant $\beta_p = 1.351$ between average avalanche size $<S>$ and duration $D$. An observed value $\beta_o = 1.351$ is obtained through regression on the avalanche data. The predicted and observed values for the critical exponent $\beta$, a measure of the criticality of a system, agree reasonably well with a difference of $|\beta_o-\beta_p| = .006$, indicating that the spike trains exhibit criticality. This is unsurprising given that the PIF's weight matrix had its maximum eigenvalue tuned to 1 (Methods). (\textbf{B}) The outputs of a naïve network with \textit{uniform} initial network characteristics failed to exhibit criticality, with both size and duration distributions deviating significantly from power laws. (\textbf{C}) After optimization, the size and duration distributions follow power laws with impressive agreement between predicted and actual critical exponents $|\beta_o-\beta_p| = .064$. (\textbf{D}) The control condition, however, also demonstrates a reasonable approximation of criticality albeit with larger disagreement between predicted and fit $\beta$: $|\beta_o-\beta_p| = .128$. This suggests that synaptic scaling without neuron-by-neuron precision weight changes may be sufficient to generate criticality in an LIF network.}
	\label{criticality}
\end{figure}

We investigated the avalanche statistics for the critical PIF reference network and compared it with the avalanche statistics for the naive, optimized, and control LIF network from ten runs of the algorithm (Figure 8). As expected, the critical PIF network demonstrates the marked agreement between the predicted and fit $\beta$ values with absolute difference $|\beta_o-\beta_p|=.006$, indicating critical avalanche statistics. Because the PIF network was already tuned to criticality, this agreement is unsurprising. In contrast, the naive LIF networks with weights drawn from a uniform distribution on $[0,1 \times 10^{-3}]$ had fewer avalanches and the avalanches that were observed failed to follow a power law with size or duration. After running our algorithm, the concatenated output spike trains of the optimized and control LIF networks were also subjected to avalanche analysis. Of the three LIF networks, the optimized network demonstrated the best agreement between its predicted and observed values of $\beta$ with absolute difference $|\beta_o-\beta_p|=.064$ as is evident in Figure \ref{criticality} (\textbf{C}). This indicates that the weight alterations made during the application of PSPM successfully induced critical activity in the optimized LIF network. Interestingly, however, the control condition, shown in Figure \ref{criticality} (\textbf{D}) exhibits decent agreement between the predicted and observed $\beta$ values with $|\beta_o-\beta_p|=.128$. This surprising agreement is of potential theoretical interest, as it could indicate that non-targeted alterations to the weight matrix of a sort similar to that of biological synaptic scaling are sufficient to induce criticality, providing a potential explanation of the emergence of criticality in biological neural networks which should be explored in future work.\\


\section*{Discussion}

We developed flexible, supervised learning rules for SNNs that reproduce a desired set of spike trains. The principle of our learning rules is to produce spike pairs, which consist of a spike in the reference outputs and a spike in the model outputs that have a bounded temporal distance. During training, spike pairs are identified with a dynamic program, and stochastic weight changes are made to eliminate or induce spikes in the model outputs to minimize the number of unpaired spikes. Our learning rules, while simple, do not require information about post-synpatic potentials or demand the network have a feedforward structure. Algorithms based on gradient descent commonly need the former while algorithms specifically performing backpropagation require the latter as well. 

In addition, we explore how training all-to-all SNNs with PSPM for spike train similarity allows the discovery of weight parameters. By comparing our learned weight matrices with those of a ground truth model, we find that spiking neural networks, and thus potentially biological neural networks, have highly degenerate connectivity. Lastly, we explore the relative contributions of local and homeostatic weight updates in this supervised learning setting.

PSPM improved spike train and inter-spike interval similarity with the desired set of spike trains. However, we find that training a model that reproduces spike trains does not entail that the resulting weight matrix matches that of the reference model. The fact that weight matrices were under-determined by the output spike trains of our model networks could have implications for connectome inference from spike data. Fitting SNN models with a high degree of recurrence to spike train data may require additional constraints on overall connectivity statistics such as sparsity, network in-degree and out-degree, as well as other network measures. Perhaps promoting sparsity during training with regularization penalties could improve the agreement between the model and the ground truth weight matrices. Given that many weight matrices can generate the same outputs \cite{caudill}, it would also be worth exploring whether incorporating data sets with a variety of external inputs into training and imposing constraints on weight matrix conectivity statistics, better agreement between resulting weight matrices may be possible. 

We did not benchmark our learning rules against other methods since we were primarily interested in whether simple learning rules of this kind could learn the weights of an SNN, especially in the case where the ground truth model is highly recurrent. Investigating the accuracy of learned weights for different learning algorithms would be an interesting direction of future research.

In addition to studying supervised learning for SNNs in the context of connectome inference, we also analyzed the relative contribution of local learning rules and network-wide homeostatic weight updates. We find that PSPM local learning rules are responsible for dramatic improvements in spike train similarity during learning, while random homeostatic adjustments are insufficient to reproduce desired spike trains. However, dramatic changes in avalanche statistics can occur solely through homeostatic changes. 

Several previous strategies for supervised learning in SNNs employed gradient based methods, which require information about the neuron model and its post-synaptic potentials \cite{bohte} \cite{booij} \cite{GHOSHDASTIDAR20091419} \cite{schrauwen1} \cite{schrauwen2}. Weight updates derived in the context of stochastic model neurons can similarly be derived from gradient descent rules \cite{Pfister} \cite{russell} \cite{gardner}. These targeted gradient-based strategies can be contrasted with evolutionary search optimization of SNNs, which only require a loss function that can be evaluated at each iteration \cite{jin} \cite{pavlidis}. The strategy of PSPM is a compromise between these competing views. The magnitude of the updates are still stochastic, like in the case of evolutionary search, but the synapses are chosen on the basis of which neurons fired in the recent past of the spike of interest. In addition, many of the previous studies on SNNs trained only one neuron that received many pre-synaptic inputs, motivating our study of learning weights that produce spike trains for an entire network \cite{Florian} \cite{Ponulak} \cite{Mohemmed}.

A limitation of PSPM is the presence of two free hyper-parameters involved in the learning process. The maximum number of timesteps that can separate two paired spikes is set by the user and is represented by the parameter $a_{cap}$. Likewise, the number of timesteps in the recent past that should be considered when making synaptic updates is also set externally with the parameter $z$. Although we do not require knowledge of the pre-synaptic potential, the choice of $z$ requires an a priori estimate of the amount of time in the past that is relevant to the production or elimination of a given spike.

Another limitation of the present work is that each of our networks were simulated with currents drawn from a single distribution and were trained to generate optimized outputs $\mathcal{O}$ as similar as possible to reference outputs $\mathcal{R}$ when exposed to identical inputs. Further work could elucidate the performance of PSPM when a network is trained on external input currents drawn from a variety of distributions.

Further empirical work is required to benchmark the performance of various SNN learning algorithms for connectome inference in highly recurrent networks. Assessing the differences between the generative models produced during training for these various algorithms would also be worth exploring.

Another interesting avenue of work would be to assess the hypothesis that PSPM is generating high-dimensional attractors in the network dynamics. As discussed in the introduction, this issue has not been addressed in the present paper, but is worthy of further analysis. Indeed, it would be interesting to demonstrate that PSPM or similar homeostatics-inspired processes are capable of producing dynamical attractors, as given the reasonable bio-realism of our neural network mode this could suggest a mechanism by which biological neural networks generate their own attractors. \cite{cabessa}\cite{cabessa2}\cite{kobayashi}\cite{asai}

In addition, future efforts may produce algorithms better suited to the above problems. Problems related to features of the network structure itself may be of interest, including the development of algorithms for replicating statistical properties of a reference network given only the inputs and spiking output of that network. Such features include in-degree and out-degree distributions, clustering coefficients, and the size distribution and number of cliques present in the network. \\

\section*{Acknowledgements}
This research was supported by a Whitehall Foundation grant (no.20121221) and a National Science Foundation Collaborative Research in Computational Neuroscience grant (no. 1308159).

\nolinenumbers
\pagebreak

\end{document}